\documentclass[10pt,twocolumn,letterpaper]{article}

\usepackage{cvpr}
\usepackage{times}
\usepackage{epsfig}
\usepackage{graphicx}
\usepackage{amsmath}
\usepackage{amssymb}
\usepackage{float}
\usepackage{url}
\usepackage{booktabs}
\usepackage{multirow}
\usepackage{subfigure}
\usepackage{makecell}
\usepackage{booktabs}

\usepackage[pagebackref=true,breaklinks=true,letterpaper=true,colorlinks,bookmarks=false]{hyperref}

\cvprfinalcopy 

\def\cvprPaperID{2026} 

\newcolumntype{x}{>\small c}
\newcolumntype{L}[1]{>{\raggedright\let\newline\\\arraybackslash\hspace{0pt}}m{#1}}
\newcolumntype{C}[1]{>{\centering\let\newline\\\arraybackslash\hspace{0pt}}m{#1}}
\newcolumntype{R}[1]{>{\raggedleft\let\newline\\\arraybackslash\hspace{0pt}}m{#1}}

\ifcvprfinal\fi
\begin{document}

\title{CRAFT Objects from Images}

\author{Bin Yang$^1$\quad\quad Junjie Yan$^2$\quad\quad Zhen Lei$^1$\thanks{Corresponding author.}\quad\quad Stan Z. Li$^1$\\
$^1$National Laboratory of Pattern Recognition\\
Institute of Automation, Chinese Academy of Sciences\\
$^2$Tsinghua University\\
{\tt\small \{bin.yang, zlei, szli\}@nlpr.ia.ac.cn\quad\quad yanjunjie@outlook.com}\\
}

\maketitle

\begin{abstract}

Object detection is a fundamental problem in image understanding. One popular solution is the R-CNN framework~\cite{girshick2014rcnn} and its fast versions~\cite{girshick2015fast,ren2015faster}. They decompose the object detection problem into two cascaded easier tasks: 1) generating object proposals from images, 2) classifying proposals into various object categories. Despite that we are handling with two relatively easier tasks, they are not solved perfectly and there's still room for improvement.

In this paper, we push the ``divide and conquer'' solution even further by dividing each task into two sub-tasks. We call the proposed method ``CRAFT'' (Cascade Region-proposal-network And FasT-rcnn), which tackles each task with a carefully designed network cascade. We show that the cascade structure helps in both tasks: in proposal generation, it provides more compact and better localized object proposals; in object classification, it reduces false positives (mainly between ambiguous categories) by capturing both inter- and intra-category variances. CRAFT achieves consistent and considerable improvement over the state-of-the-art on object detection benchmarks like PASCAL VOC 07/12 and ILSVRC.

\end{abstract}

\section{Introduction}

\begin{figure}[t]
\begin{center}
   \includegraphics[width=0.95\linewidth]{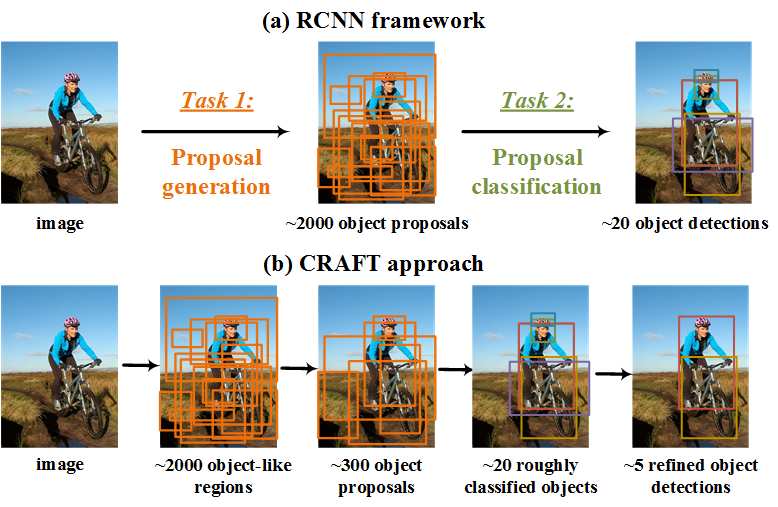}
\end{center}
   \caption{Overview of the widely used two-step framework in object detection, and the proposed CRAFT pipeline.}
\label{fig:pipeline}
\end{figure}

The problem definition of object detection is to determine where in the image the objects are and which category each object belongs to. The above definition gives us a clue of how to solve such a problem: by generating object proposals from an image (where they are), and then classifying each proposal into different object categories (which category it belongs to). This two-step solution matches to some extent with the attentional mechanism of humans seeing things, which is to first give a coarse scan of the whole scenario and then focus on regions of our interest. 

As a matter of fact, the above intuitive solution is where the research community is moving forward for years. Recently, the two steps (proposal generation and object classification) have been solved quite satisfactorily by two advances in computer vision: first is the introduction of general object proposals, second is the revival of the Convolutional Neural Networks (CNN). The general object proposal algorithm (e.g., Selective Search \cite{uijlings2013selective} and EdgeBox \cite{zitnick2014edge}) can provide around 2000 proposals per image to cover most of the objects and make the employment of more complex classifier for each proposal possible. The prosperity of the Convolutional Neural Networks (CNN) comes from its rich representation capacity and powerful generalization ability in image recognition, which is proved in challenging ImageNet classification task \cite{krizhevsky2012imagenet,szegedy2015going,simonyan2014very}. With the off-the-shelf methods available, the seminal work R-CNN \cite{girshick2014rcnn} shows that Selective Search based region proposals plus the CNN based object classifier can achieve very promising performance in object detection. The R-CNN framework is further improved by Fast R-CNN \cite{girshick2015fast} and Faster R-CNN \cite{ren2015faster}, while the former enables end-to-end learning of the whole pipeline, and the latter introduces the Region Proposal Network (RPN) to get object proposals of higher quality.

Although the R-CNN framework achieves superior performance on benchmarks like PASCAL VOC, we discover quite large room for improvement after a detailed analysis of the result on each task (proposal generation and classification). We claim that there exists an offset between current solution and the task requirement, which is the core problem of the popular two-step framework. Specifically, in proposal generation, the task demands for proposals for only objects, but the output of general object proposal algorithms still contains a large proportion of background regions. In object classification, the task requires classification among objects, while practically in R-CNN it becomes classification among object categories plus background. The existence of many background samples makes the feature representation capture less intra-category variance and more inter-category variance (ie, mostly between the object category and background), causing many false positives between ambiguous object categories (eg, classify tree as potted plant). 

Inspired by the ``divide and conquer'' strategy, we propose to further divide each task via a network cascade to alleviate the above issues (see Figure \ref{fig:pipeline} for an illustration). Practically, in proposal generation task, we add another CNN based classifier to distinguish objects from background given the output of off-the-shelf proposal algorithm (eg, Region Proposal Network); and in object classification task, since the N+1 class (N object categories plus background) cross-entropy objective leads the feature representation to learn inter-category variance mainly, we add a binary classifier for each object category in order to focus more on intra-category variance. Through delicate design of the cascade structure in each task, we discover that it helps a lot: object proposals are more compact and better localized, while the detections are more accurate with fewer false positives between ambiguous object categories. 

As a result, the object detection performance gets improved by a large margin. We show consistent and considerable gain over the Faster R-CNN baseline in object detection benchmark PASCAL VOC 07/12 as well as the more challenging ILSVRC benchmark.

The remainder of the paper is organized as follows. We review and analyze related works in Section 2. Our CRAFT approach is illustrated in Section 3 and validated in Section 4 respectively. Section 5 concludes the paper.

\section{Related work}

CRAFT can be seen as an incremental work built upon the state-of-the-art two-step object detection framework. In order to give readers a full understanding of our work and the underlying motivation, in this section we first review the development of the two-step framework from the ``divide and conquer'' perspective. We introduce in turn the significant advances in proposal generation and object classification respectively. After a summary of the building stones, we briefly introduce some related works that also try to improve upon the state-of-the-art two-step framework and also show our connection with them.

\subsection{Development of the two-step framework}

Proposals are quite important for object detection and diverse methods for object proposal generation are proposed. In case of detecting one particular category of near rigid objects (like faces or pedestrians) with fixed aspect ratio, sliding window mechanism is often used \cite{oren1997pedestrian,rowley1998neural,viola2004robust}. The main disadvantage is that the number of candidate windows can be about $O(10^6)$ for an image, therefore limiting the complexity of classifier due to efficiency issues. When it comes to generating proposals covering general objects of various categories and in various shapes, sliding window approach becomes more computationally expensive.

Many works are proposed to get more compact proposals, which can be divided into two types: the unsupervised grouping style and the supervised classification style. The most popular method in grouping style is the Selective Search \cite{uijlings2013selective}, which hierarchically groups super-pixels generated through \cite{felzenszwalb2004efficient} to form general object proposals. Other typical grouping style proposal methods include the EdgeBox \cite{zitnick2014edge} which is faster and MCG \cite{arbelaez2014multiscale} which is more compact. With around 2000 proposals kept for each image, a recall rate of $98\%$ on Pascal VOC and 92\% on ImageNet can be achieved. Besides the smaller number of proposals, another advantage of grouping style over sliding window is that proposals at arbitrary scale and aspect ratio can be generated, which provides much more flexibility. Many works have been proposed for further improvement and an evaluation can be found in \cite{Hosang2014Bmvc}. 

In the supervised camp, the proposal generation problem is defined as a classification and/or regression problem. Typical methods include the BING \cite{BingObj2014} and Multi-box \cite{szegedy2013deep,erhan2013scalable}. The BING uses the binary feature and SVM to efficiently classify objects from background. The Multi-box uses CNN to regress the object location in an end-to-end manner. A recently proposed promising solution is the Region Proposal Network (RPN) \cite{ren2015faster}, where a multi-task fully convolutional network is used to jointly estimate proposal location and assign each proposal with a confidence score. The number of proposals is also reduced to be less than 300 with higher recall rate. We use the RPN as the baseline proposal algorithm in CRAFT.

Given object proposals, detection problem becomes an object classification task, which involves representation and classification. Browsing the history of computer vision, the feature representation is becoming more and more sophisticated, from hand-craft Haar \cite{viola2004robust} and HOG \cite{dalal2005histograms} to learning based CNN \cite{girshick2014rcnn}. Built on top of these feature representations, carefully designed models can be incorporated. The two popular models are the Deformable Part Model (DPM \cite{felzenszwalb2010object}) and the Bag of Words (BOW \cite{perronnin2010improving,Chatfield11}). Given the feature representation, classifiers such as Boosting \cite{friedman2000additive} and SVM \cite{cortes1995support} are commonly used. Structural SVM \cite{tsochantaridis2005large,Joachims/etal/09a} and its latent version \cite{yu2009learning} are widely used when the problem has a structural loss.

In recent three years, with the revival of CNN \cite{krizhevsky2012imagenet}, CNN based representation achieves excellent performance in various computer vision tasks, including object recognition and detection. Current state-of-the-art is the R-CNN approach. The Region-CNN (R-CNN) \cite{girshick2014rcnn} is the first to show that Selective Search region proposal and the CNN together can produce a large performance gain, where the CNN is pre-trained on large-scale datasets such as ImageNet to get robust feature representation and fine-tuned on target detection dataset. Fast R-CNN \cite{girshick2015fast} improves the speed by sharing convolutions among different proposals \cite{kaiming14ECCV} and boosts the performance by multi-task loss (region classification and box regression). \cite{ren2015faster} uses Region Proposal Network to directly predict the proposals and makes the whole pipeline even faster by sharing full-image convolutional features with the detection network. We use the Fast R-CNN as the baseline object classification model in CRAFT.

\subsection{Improvements on the two-step framework}

Based on the two-step object detection framework, many works have been proposed to improve it. Some of them focus on the proposal part. \cite{ouyang2014deepid,yan2015object} find that using the CNN to shrink the proposals generated by grouping style proposals leads to performance gain. \cite{ghodrati2015deepboxes,kuo2015deepbox} use CNN cascade to rank sliding windows or re-rank object proposals. CRAFT shares both similarities and differences with these methods. The common part is that we both the ``cascade'' strategy to further shrink the number of proposals and improve the proposal quality. The discrepancy is that those methods are based on sliding window or grouping style proposals, while ours is based on RPN which already has proposals of much better quality. We also show that RPN proposals and grouping style proposals are somewhat complementary to each other and they can be combined through our cascade structure. 

Some other works put the efforts in improving the detection network (R-CNN and Fast R-CNN are popular choices). \cite{gidaris2015object} proposes the multi-region pipeline to capture fine-grained object representation. \cite{bell15ion} introduces the Inside-Outside Net, which captures multi-scale representation by skip connections and incorporates image context via spatial recurrent units. These works can be regarded as learning better representation, while the learning objective is not changed. In CRAFT, we identify that current objective function in Fast R-CNN leads to flaws in the final detections, and address this by cascading another complementary objective function. In other words, works like \cite{gidaris2015object,bell15ion} that aim to learn better representation are orthogonal to our work.

In a word, guided by the ``divide and conquer'' philosophy, we propose to further divide the two steps in current state-of-the-art object detection framework, and both tasks are improved considerably via a delicate design of network cascade. Our work is complementary to many other related works as well. Besides these improvements built on the two-step framework, there are also some works \cite{lenc2015r,stewart2015end,redmon2015you} on end-to-end detection framework that drops the proposal step. However, these methods work well under some constrained scenarios but the performance
drops notably in general object detection in unconstrained environment.

\section{The CRAFT approach}

In this section we explain why we propose CRAFT, how we design it and how it works. Following the proposal generation and classification framework, we elaborate in turn how we design the cascade structure based on the state-of-the-art solutions to solve each task better. Implementation details are presented as well.

\subsection{Cascade proposal generation}

\subsubsection{Baseline RPN}
An ideal proposal generator should generate as few proposals as possible while covering almost all object instances. With the help of strong abstraction ability of CNN deep feature hierarchies, RPN is able to capture similarities among diverse objects. However, when classifying regions, it is actually learning the appearance pattern of an object that distinguishes it from non-object (such patterns may be colorful segments, sharp and closed edges). Therefore its outputs are actually object-like regions. The gap between object-like regions and the demanded output -- object instances -- makes room for improvement. In addition, due to the resolution loss caused by CNN pooling operation and the fixed aspect ratio of sliding window, RPN is weak at covering objects with extreme scales or shapes. On the contrast, the grouping style methods are complementary in this aspect. 

To analyze the performance of the RPN method, we train a RPN model based on the VGG\_M model (defined in \cite{simonyan2014very}) using PASCAL VOC 2007 train+val and show its performance in Table~\ref{tab:recallablation}. The recall rates in the table are calculated with 0.5 IoU (intersection of union) criterion and 300 proposals per image on the PASCAL VOC 2007 test set. The overall recall rate of all object categories is 94.87\%, but the recall rate on each object category varies a lot. In accordance with our assumption, objects with extreme aspect ratio and scale are hard to be detected, such as boat and bottle. What's more, objects with less appearance complexity, or those usually immersed in object clutters, are also difficult to be distinguished from background by RPN, like plant, tv and chair.

\begin{table}
\begin{center}
\begin{tabular}{|c|c|c|c|c|}
\hline
aero      & bike      & \bf{bird}      & \bf{boat}      & \bf{bottle} \\
\hline
95.44 & 98.81 & \bf{93.90} & \bf{92.78} & \bf{80.38} \\
\hline\hline
bus        & car       & cat        & \bf{chair}      & cow  \\
\hline
98.12 & 96.00 & 99.16 & \bf{91.80} & 99.18 \\
\hline\hline
table      & dog        & horse  & mbike & persn \\
\hline
95.15 & 99.59 & 97.70 & 96.31 & 95.49 \\
\hline\hline
\bf{plant}      & sheep      & sofa       & train    & \bf{tv}  \\
\hline
\bf{86.87} & 98.76 & 98.74 & 97.52 & \bf{90.58} \\
\hline
\end{tabular}
\end{center}
\caption{Recall rates (\%) of different classes of objects on VOC2007 test set, using 300 proposals from a Region Proposal Network for each image. The overall recall rate is 94.87\%, and categories that get lower recall rates are highlighted. VGG\_M model is used as network initialization.}
\label{tab:recallablation}
\end{table}

\subsubsection{Cascade structure}
\begin{figure}[t]
\begin{center}
   \includegraphics[width=0.95\linewidth]{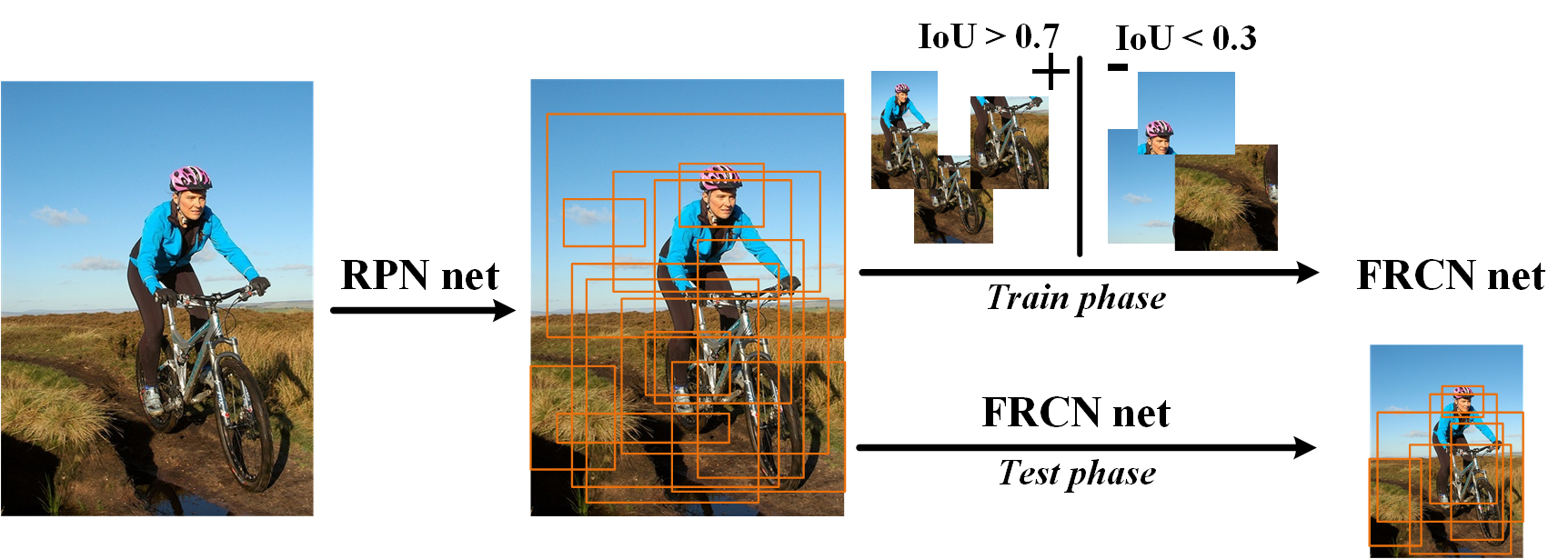}
\end{center}
   \caption{The pipeline of the cascade proposal generator. We first train a standard Region Proposal Network (RPN net) and then use its output to train another two-class Fast-RCNN network (FRCN net). During testing phase, the RPN net and the FRCN net are concatenated together. The two nets do not share weights and are trained separately from the same pre-trained model.}
\label{fig:crpn}
\end{figure}

In order to make a bridge between the object-like regions provided by RPN and the object proposals demanded by the detection task, we introduce an additional classification network that comes after the RPN. According to definition, what we need here is to classify the object-like regions between real object instances and background/badly located proposals. Therefore we take the additional network as a 2-class detection network (denoted as FRCN net in Figure~\ref{fig:crpn}) which uses the output of RPN as training data. In such a cascade structure, the RPN net takes universal image patches as input and is responsible to capture general patterns like texture, while the FRCN net takes input as object-like regions, and plays the role of learning patterns of finer details. 

The advantages of the cascade structure are two-fold: First, the additional FRCN net further improves the quality of the object proposals and shrinks more background regions, making the proposals fit better with the task requirement. Second, proposals from multiple sources can be merged as the input of FRCN net so that complementary information can be used.

\subsubsection{Implementation}
We train the RPN and FRCN nets consecutively. The RPN net is trained regularly in a sliding window manner to classify all regions at various scales and aspect ratios in the image, with the same parameters as in \cite{ren2015faster}. After the RPN net is trained, we test it on the whole training set to produce 2000 primitive proposals of each training image. These proposals are used as training data to train the binary classifier FRCN net. Note that when training the second FRCN net, we use the same criterion of positive and negative sampling as in RPN (above 0.7 IoU for positives and below 0.3 IoU for negatives). 

At testing phase, we first run the RPN net on the image to produce 2000 primitive proposals and then run FRCN net on the same image along with 2000 RPN proposals as the input to get the final proposals. After proper suppression or thresholding, we can get fewer than 300 proposals of higher quality.

We use the FRCN net rather than RPN net as the second binary classifier for that FRCN net has more parameters in its higher-level connections, making it more capable to handle with the more difficult classification problem. If we use the model definition of RPN net as the second classifier, the performance degrades. In our current implementation, we do not share full-image convolutional features between RPN net and FRCN net. If we share them, we expect little performance gain as in \cite{ren2015faster}.

\subsection{Cascade object classification}

\subsubsection{Baseline Fast R-CNN}

A good object classifier is supposed to classify each object proposal correctly into certain number of categories. Due to the imperfection of the proposal generator, there exists quite a large number of background regions and badly located proposals in the proposals. Therefore when training the object classifier, an additional object category is often added as ``background''. In the successful solution Fast R-CNN, the classifier is learned with a multi-class cross-entropy loss through softmax layer. Aided by the auxiliary loss of bounding box regression, the detection performance is superior to ``softmax + SVM'' paradigm in R-CNN approach. In order to get an end-to-end system, Fast R-CNN drops the one-vs-rest SVM in R-CNN, which creates the gap between the resulting solution and the task demand.

\begin{figure}[t]
\begin{center}
   \includegraphics[width=0.4\linewidth]{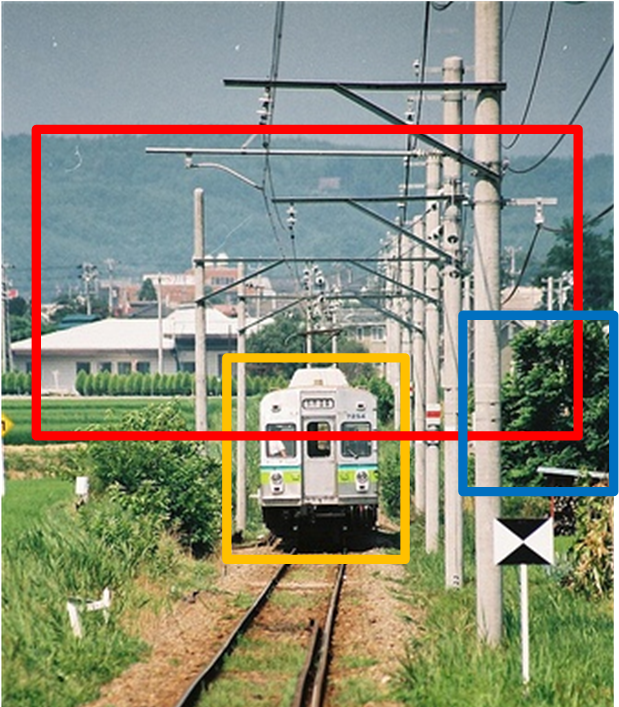}
\end{center}
   \caption{Example detections from a Fast-RCNN model. Different colors indicate different object categories. Specifically, orange color denotes ``train'', red denotes ``boat'' and blue denotes ``potted plant''.}
\label{fig:miscalss}
\end{figure}

Given object proposals as input and final object detections as output, the task demands for not only further distinguishing objects of interested categories from non-objects, but also classifying objects into different classes, especially those with similar appearance and/or belong to semantically related genres (car and bus, plant and tree). This calls for a feature representation that captures both the inter-category and intra-category variances. In the case of Fast R-CNN, the multi-class cross-entropy loss is responsible for helping the learned feature hierarchies capture inter-category variance, while it is weak at capturing intra-category variance as the ``background'' class usually occupies a large proportion of training samples. Example detection results of Fast R-CNN are shown in Figure~\ref{fig:miscalss}, where the mis-classification error is a major problem in the final detections.

\subsubsection{Cascade structure}
\begin{figure}[t]
\begin{center}
   \includegraphics[width=0.95\linewidth]{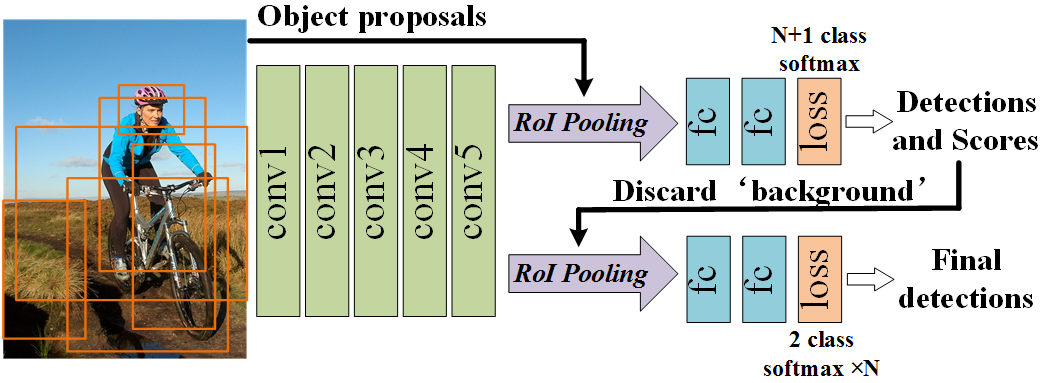}
\end{center}
   \caption{The work flow of the cascade proposal classifier. We first train a standard Fast-RCNN network (FRCN-1) and use its output scores to assign each detection with a class label. Then detections belonging to ``background'' are discarded and the rest are used to train another Fast-RCNN network (FRCN-2) whose loss is the sum of N two-class softmax losses. Note that the auxiliary bounding box regression loss is also used in both FRCN nets but left out in the figure for better presentation. The two FRCN nets are optimized consecutively with shared convolution weights so that the image feature maps are computed only once during testing phase.}
\label{fig:cfrcn}
\end{figure}

To ameliorate the problem of too many false positives caused by mis-classification, we bring the one-vs-rest classifier back in the form of an additional two-class cross-entropy loss for each object category (shown in Figure~\ref{fig:cfrcn}). In essence, the added classifier is playing the role of SVM in R-CNN framework. We find it important to train each one-vs-rest classifier using the detection output of that specific category (meaning the detection should have highest score on that specific category). In this way, each one-vs-rest classifier sees proposals specific to one particular object category (also containing some false positives), making it focused at capturing intra-category variance. 

For example, in PASCAL VOC dataset, the training samples for the additional classifier of class ``potted plants'' are usually trees, grass, potted plants and some other green things. After the training, it is able to capture the minute difference between various types of plants, so as to reduce false positives related to this class. This effect can hardly be achieved through a multi-class cross-entropy loss.

\subsubsection{Implementation}
During the training phase, a standard FRCN net (FRCN-1) is first trained using object proposals from the cascade proposal generator. Thereafter, we train another FRCN net (FRCN-2) based on the output of FRCN-1 (which we call primitive detections). Since we are now dealing with classification task among objects, we discard the primitive detections which are classified as ``background''. The objective function of the FRCN-2 is the sum of N 2-class cross-entropy losses (N equals the number of object categories), with each 2-class classifier depends only on primitive detections assigned with the corresponding class label. The criterion of positive and negative sampling for the one-vs-rest classifier is the same as RPN. Practically, there are roughly 20 primitive detections per image used for FRCN-2 training, which is quite limited. 

To effectively train FRCN-2 and efficiently detect objects from proposals, we share the convolution weights of FRCN-1 and FRCN-2 so that the full-image feature maps need only be computed once. That is to say, the convolution weights of FRCN-2 are initialized from FRCN-1 and keep fixed during FRCN-2 training. The fully-connected layers of FRCN-2 are initialized from FRCN-1 as well, and new layers to produce 2N scores and 4N bounding box regression targets are initialized from a gaussian distribution. 

At test time, with 300 object proposals as input, FRCN-1 outputs around 20 primitive detections, each with N primitive scores. Then each primitive detection is again classified by FRCN-2 and the output scores (N categories) is multiplied with the primitive scores (N categories) in a category-by-category way to get the final N scores for this detection.

\section{Experiments}

We first validate that the proposed cascade structure does improves the performance of each task in the two-step object detection framework through a delicate design, then we show the overall performance gain in object detection by evaluating CRAFT on benchmarks like PASCAL VOC 07/12 and ILSVRC. Note that we do not share full-image convolutional features between the proposal generation and classification tasks, therefore the proper baseline would be the unshared version of Faster R-CNN \cite{ren2015faster}.

\subsection{Proposal generation}

Firstly we justify the design choice of the cascade proposal generator. We answer two questions: 1) do we really need a more complex network in the second stage? 2) do we need the strict sampling criterion during training? We show evaluation of different parameterization in Table \ref{tab:crpndesign}. Since RPN already performs quite well on PASCAL VOC benchmarks, we show parameterization evaluation on the more challenging ILSVRC dataset, and then present a thorough evaluation of the final design of the cascade proposal generator on PASCAL VOC.

We show comparison of different choices of the sampling criterion and network definition of the cascade binary classifier in Table \ref{tab:crpndesign}. All models in the table are initialized from a pre-trained VGG19 model, trained on the ILSVRC DET train+val1 sets and tested on val2 set\footnote{The splits of ``val1'' and ``val2'' are the same as \cite{girshick2014rcnn}}, with a evaluation metric of 0.5IoU threshold. To handle with small objects in ILSVRC, we add two additional anchor scales (64 and 32) in RPN and change the batch-size to 2 images. However, the RPN's performance (89.94\%) is still inferior to Selective Search (92.09\%). When cascaded with an additional binary classifier (``+FRCN''), the recall rate increases by over 2\%. 

We show that the strict sampling criterion (0.7IoU threshold for positives, 0.3IoU threshold for negatives) leads to slightly better performance. When we replace the FRCN with a RPN-like network definition which uses a 512-d feature representation rather than 4096-d, the recall rate degrades. With the best design choice, we can achieve higher recall rate (92.37\%) than Selective Search with only 300 proposals.

\begin{table}
\begin{center}
\begin{tabular}{c|c|c|c}
Model      & IoUthr\_pos   &  IoUthr\_neg     & Recall (\%) \\
\hline
SS & - & - & 92.09 \\
RPN & 0.7 & 0.3 & 89.94 \\
\hline
+ FRCN & 0.5 & 0.5 & 92.13\\
+ FRCN & 0.5 & 0.3 & 92.24 \\
+ FRCN & 0.7 & 0.3 & \bf{92.37}  \\
\hline
+ RPN-2  & 0.7 & 0.3 & 91.83 \\
\end{tabular}
\end{center}
\caption{Evaluation results of different design choices of the cascaded binary classifier in cascade proposal generator on ILSVRC DET val2 set. All recall rates except that of Selective Search (``SS'') are reported with 300 proposals per image. For Selective Search baseline, there are roughly 2000 proposals per image.}
\label{tab:crpndesign}
\end{table}

Next we throughly evaluate our cascade proposal generator on PASCAL VOC in Table \ref{tab:box}. Baselines are Selective Search (``SS''), RPN (VGG16 net with 512-d feature representation), and RPN\_L (VGG16 net with 4096-4096-d feature representation, meaning larger RPN). We evaluate with regard to not only recall rates at different IoU thresholds (from 0.5 to 0.9), but also the mAP of a Fast R-CNN detector trained on different proposal algorithms, which makes the comparison more meaningful because the proposals are eventually used for object detection. Note than all the methods in Table \ref{tab:box} are purely for proposal task without joint optimization with object detection.

\begin{table}[H]
\begin{center}
\begin{tabular}{x|x|xxxxx|x}
method & \#box & 0.5 & 0.6 & 0.7 & 0.8 & 0.9 & mAP\\
\hline
SS & 2000 & 92.1 & 85.2 & 72.5 & 52.9 & \bf{26.6} & 70.0\\
\hline
RPN & 2000 & \bf{98.5} & \bf{95.8} & 84.1 & 40.7 & 4.1 & -\\
RPN & 300 & 96.3 & 92.5 & 78.8 & 37.9 & 3.9 & 71.6\\
RPN\_L & 300 & 95.4 & 90.3 & 76.5 & 37.4 & 3.8 & -\\
\hline
Ours & 300 & 97.9 & 95.5 & \bf{89.6} & \bf{63.7} & 13.0 & 72.2 \\
Ours\_S & \bf{87} & 96.8 & 94.1 & 87.8 & 62.4 & 12.9 & \bf{72.5}\\
\end{tabular}
\end{center}
\caption{Proposal evaluation by recall rate (\%) with regard to different IoUs and detection mAP (\%) on PASCAL VOC 07 test set. All CNN based methods use VGG16 net as initialization and are trained on PASCAL VOC 07+12 trainval set. ``Ours'' keeps fixed number of proposals per image (same as RPN), while ``Ours\_S'' keeps proposals whose scores (output of the cascaded FRCN classifier) are above a fixed threshold.}
\label{tab:box}
\end{table}

From the table we can see that: (1) RPN proposals aren't so well localized compared with bottom-up methods (low recall rates at high IoU thresholds). (2) This cannot be ameliorated by using a larger network because it is caused by fixed anchors. (3) Our cascaded proposal generator not only further eliminates background proposals, but also brings better localization, both help in detection AP.

\subsection{Object classification}

In this part we justify the use of the one-vs-rest classifier as well as explain how many layers to fine-tune for the one-vs-rest classifier. We show the evaluation results in Table \ref{tab:ensemble}. 

The cascade object classifier can be regarded as a concatenation of two FRCN nets (FRCN-1 + FRCN-2). In the top table in Table \ref{tab:ensemble} we compare several strategies concerning training the FRCN-2 net, which are no fine-tuning (``the same''), fine-tuning the additional one-vs-rest classifier weights (``cls''), fine-tuning layers above the last convolution maps (``fc + clf''), and fine-tuning all layers except for conv1 (``conv + fc + clf''). In fact, ``the same'' is simply running the FRCN-1 net twice, and hopefully the iterative bounding box regression would help improve the result. Another three fine-tuning strategies are trying to improve the performance by introducing additional class-specific one-vs-rest classifiers to capture intra-category variance. The difference between these three is different level of feature sharing with FRCN-1 net: ``clf'' uses exactly the same feature representation as FRCN-1, and ``conv + fc + clf'' trains totally new feature representation for itself. 

From the results in the table we can see that iteratively detecting twice improves the results a little, which mainly comes from iterative bounding box regression. As for fine-tuning the net with a binary softmax loss, different settings vary in performance. In a word, through sharing convolutional features but fine-tuning high-level connections we get best result. There are two possible reasons that account for it: 1) the training samples for the FRCN-2 are limited and biased; 2) we just want to learn another classifier rather than learn total different feature representation. What's more, the improvement gained by the cascade approach is more than that by iteratively detecting twice, proving that the one-vs-rest softmax loss does play part of the role of hard negative mining and helps reduce the mis-classification error.

We additionally justify the cascade structure in the bottom table in Table \ref{tab:ensemble}. One-vs-rest classifier alone performs poorly because each binary classifier has to handle with objects of various classes but binary label provides limited information, while in our case each binary classifier only handles with detections of one class (ie, detection output of the FRCN-1 net), making it more specialized.

\begin{table}
\begin{center}
\begin{tabular}{c|c|c|c}
FRCN-1  & FRCN-2  &  FT layers & mAP (\%) \\
\hline
VGG\_M & - & - & 65.0 \\
VGG\_M & the same & - & 65.2 \\
\hline
VGG\_M & VGG\_M & clf & 66.3\\
VGG\_M & VGG\_M & fc + clf & \bf{68.0}\\
VGG\_M & VGG\_M & conv + fc + clf  & 67.7  \\
\end{tabular}
\end{center}

\begin{center}
\begin{tabular}{x|x}
classifier objective & mAP (\%) \\
\hline
FRCN (one-shot) & 65.0 \\
one-vs-rest & 46.1 \\
\hline
Ours (one-shot + one-vs-rest) & \bf{68.0} \\
\end{tabular}
\end{center}
\caption{\textbf{Top:} Evaluation of how many layers to fine-tune for the one-vs-rest classifier. \textbf{Bottom:} Evaluation of different classifier objectives. All models use VGG\_M as network initialization. All results are evaluated on PASCAL VOC 07 (trainval for training, test for testing) with the same object proposals from a trained RPN model.}
\label{tab:ensemble}
\end{table}

\subsection{Object detection}

After showing the superiority of CRAFT on both tasks in the two-step object detection framework, we now evaluate CRAFT on object detection benchmarks. We first evaluate CRAFT on PASCAL VOC 07\&12 in comparison with the state-of-the-art detectors Fast R-CNN and Faster R-CNN, and then show our results on the more challenging ILSVRC benchmark.

\subsubsection{PASCAL VOC 2007 \& 2012}

\begin{figure}[t]
\begin{center}
   \includegraphics[width=0.95\linewidth]{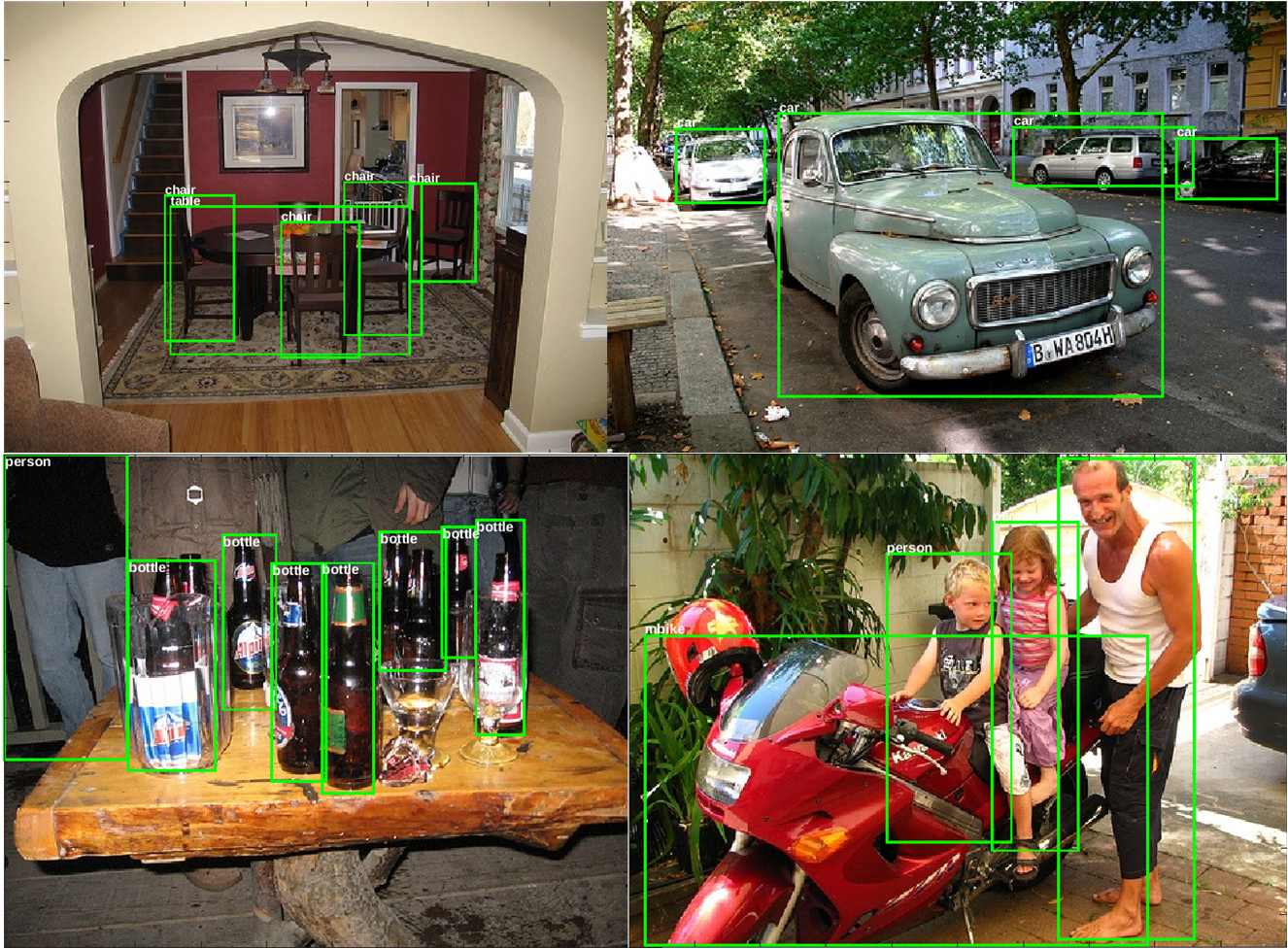}
\end{center}
   \caption{Example detections of CRAFT on PASCAL VOC 2007 test set.}
\label{fig:ex}
\end{figure}

We compare CRAFT with state-of-the-art detectors under the two-step detection framework, which are Fast R-CNN \cite{girshick2015fast} and Faster R-CNN \cite{ren2015faster}. The comparative results on PASCAL VOC 2007 \& 2012 are shown in Table \ref{tab:det}. Qualitative results on PASCAL VOC 2007 test set are shown in Figure \ref{fig:ex}. All methods use VGG16 model, and ``RPN\_un'' represents the unshared version of Faster R-CNN. All baseline results are got from original papers or by running the original open source codes. On PASCAL VOC 2007, all methods use 07+12 trainval as training data. CRAFT outperforms the baseline ``RPN\_un'' by 4.1\% absolute value in mAP (from 71.6\% to 75.7\%). On PASCAL VOC 2012, all methods use 12 trainval as training data, and this time CRAFT achieves an edge of 5.8\% absolute value in mAP (from 65.5\% to 71.3\%).

We do not compare with many other detectors which also improve over the basic two-step detection framework like \cite{gidaris2015object,bell15ion} because we believe that their contributions are orthogonal to ours. If we incorporate their methods in CRAFT, as well as using end-to-end multi-task network cascade training \cite{dai2015instance}, we expect notable further improvement.

\begin{table}[H]
\begin{center}
\begin{tabular}{x|x|x|xx}
method & proposal & classifier & voc07 & voc12 \\
\hline
FRCN \cite{girshick2015fast} & SS & FRCN & 70.0 & 65.7 \\
RPN\_un \cite{ren2015faster} & RPN & FRCN & 71.6 & 65.5$^\dagger$ \\
RPN \cite{ren2015faster} & RPN & FRCN & 73.2 & 67.0 \\
\hline
CRAFT & cascade & FRCN & 72.5 & - \\
CRAFT & cascade & cascade & \bf{75.7} & \bf{71.3}$^\ddagger$ \\
\end{tabular}
\end{center}
\caption{Object detection mAP (\%) on PASCAL VOC 07+12. ``voc07'': 07+12 trainval for training, VGG16 net. ``voc12'': 12 trainval for training, VGG16 net. ``FRCN'' and ``RPN'' results are from original paper and report. ``RPN\_un'' are Faster R-CNN with unshared feature, whose results are got from open source codes (the proper baseline). Joint optimization is not used in ``CRAFT'', which would otherwise bring some gain. $^\dagger$: \footnotesize{\url{http://host.robots.ox.ac.uk:8080/anonymous/AITNWY.html}}, $^\ddagger$: \footnotesize{\url{http://host.robots.ox.ac.uk:8080/anonymous/FFJGZH.html}}}
\label{tab:det}
\end{table}

\subsubsection{ILSVRC object detection task}

We validate that CRAFT generalizes well to large-scale problems like 200-class ILSVRC object detection task.

As shown in Table \ref{tab:crpndesign}, RPN does not generalize very well to large-scale object detection tasks even if more scales are added to the anchors. However, with the help of our cascade structure, the recall rates boosts to be over Selective Search. However, the performance is still inferior to that on PASCAL VOC. Therefore, we add additional some additional modules to the cascade proposal generator to further improve its performance. 

As shown in Table \ref{tab:ilsvrc} top, using a stricter NMS policy (0.6 IoU threshold) increases the recall rate a bit because the localization accuracy of proposals has already been improved after the cascade structure. Re-scoring each proposal by considering both scores from two stages of cascade structure also helps. Finally, fusion of multiple proposal sources boosts the recall rate to be over 94\%. We combine proposals output from ``DeepBox'' \cite{kuo2015deepbox} or ``SS'' (Selective Search) with the RPN proposals as the fusion input to the FRCN net in the cascade structure. Results show that ``DeepBox'' is better than ``SS''.

\begin{table}[H]
\begin{center}
\begin{tabular}{c|c|c|cc}
Basic & 0.6 NMS  &  re-score  & +DeepBox & +SS \\
\hline
92.37 & 93.61 & 93.75 & \bf{94.13} & 93.04  \\
\end{tabular}
\end{center}

\begin{center}
\begin{tabular}{x|x|x|x}
method & proposal & classifier & ilsvrc\\
\hline
Ouyang \etal \cite{ouyang2014deepid} & SS+EB & RCNN & 45.0 \\
Yan \etal \cite{yan2015object} & SS+EB & RCNN & 45.4 \\
\hline
RPN\_un \cite{ren2015faster} & RPN & FRCN & 45.4 \\
\hline
CRAFT & cascade & FRCN & 47.0 \\
CRAFT & cascade & cascade & \bf{48.5} \\
\end{tabular}
\end{center}

\caption{\textbf{Top:} Recall rate (\%) of cascade proposal generator on ILSVRC detection val2 set with regard to 0.5 IoU evaluation metric. \textbf{Bottom:} Detection mAP (\%) of CRAFT on ILSVRC detection val2 set in comparison with other state-of-the-art detectors.}
\label{tab:ilsvrc}
\end{table}
 
Given high-quality object proposals, we train a regular Fast R-CNN detector and a cascade object classifier upon it. We use a GoogLeNet model with batch normalization \cite{ioffe2015batch} (8.4\% top-5 validation error on ILSVRC image classification task) as network initialization. We use ILSVRC 2013train + 2014train + val1 as training set, and evaluate it on val2 set. Since 2013train set does not align well with detection task, we adopt the following batch sampling strategy: each batch is made up of 12 images, with 8 from fully annotated sets (2014train + val1) and 4 from partially annotated sets (2013train). For each fully annotated image, we sample 32 proposals with 8 positives and 24 negatives, and the IoU threshold for distinguishing positives and negatives is 0.5. For each partially annotated image, we sample 8 proposals with 2 positives and 6 negatives, and the IoU range for positives is larger than 0.7 and for negatives it is smaller than 0.5. Due to large batch-size, we train the detector on a 4-GPU implementation. 

In Table \ref{tab:ilsvrc} bottom, a regular Fast R-CNN detector achieves 47.0\% mAP on val2, which already surpasses the ensemble result of previous state-of-the-art systems like Superpixel Labeling \cite{yan2015object} and Deepid-Net \cite{ouyang2014deepid}. This edge is basically from better proposals. With cascade object classifier added, the mAP gets additional 1.5\% absolute gain.

\section{Conclusion}

In this paper, we propose the CRAFT (Cascade Region-proposal-network And FasT-rcnn) for general object detection following the ``divide and conquer'' philosophy. It improves both the proposal generation and classification tasks through carefully designed convolutional neural network cascades. For the proposal task, CRAFT outputs more compact and better localized object proposals. For detection task, CRAFT helps the network learn both inter- and intra-category variances so that false positives among ambiguous categories are largely eliminated. CRAFT achieves consistent and considerable improvements over state-of-the-art methods on PASCAL VOC and ILSVRC benchmarks, while being complementary to many other advances in object detection.

\paragraph{Acknowledgements} The authors were supported by Chinese National Natural Science Foundation Projects \#61375037, \#61473291, \#61572501, \#61502491, \#61572536, by National Science and Technology Support Program Project \#2013BAK02B01, by Chinese Academy of Sciences Project No. KGZD-EW-102-2, and by AuthenMetric R\&D Funds. We thank NVIDIA gratefully for GPU hardware donation and the reviewers for their many constructive comments.

{\small
\bibliographystyle{ieee}
\bibliography{egbib}
}

\end{document}